\documentclass[11pt]{article}

\usepackage[final]{acl}

\usepackage{times}
\usepackage{latexsym}
\usepackage{graphicx}
\usepackage{booktabs}
\usepackage{tcolorbox}
\tcbuselibrary{listings, breakable, skins}
\usepackage{float}
\usepackage{subcaption}

\newtcolorbox{promptbox}[2][]{%
  enhanced, breakable,
  colback=#2!3!white,
  colframe=#2!60!black,
  coltitle=white,
  fonttitle=\bfseries\small,
  left=5pt, right=5pt, top=4pt, bottom=4pt,
  boxrule=0.4pt,
  titlerule=0pt,
  toptitle=3pt, bottomtitle=3pt,
  colbacktitle=#2!70!black,
  arc=2pt,
  #1
}
\definecolor{promptblue}{HTML}{2563EB}
\definecolor{promptgreen}{HTML}{16A34A}
\definecolor{promptorange}{HTML}{D97706}
\definecolor{promptpurple}{HTML}{7C3AED}

\newtcbox{\codeinline}{on line, colback=gray!10, colframe=gray!10, boxrule=0pt, boxsep=0pt, left=2pt, right=2pt, top=1pt, bottom=1pt, fontupper=\ttfamily\scriptsize}

\newtcolorbox{codeblock}{
  enhanced, colback=gray!8!white, colframe=gray!30!white,
  boxrule=0.3pt, arc=1.5pt,
  left=4pt, right=4pt, top=3pt, bottom=3pt,
  fontupper=\ttfamily\scriptsize
}

\newcommand{\promptlabel}[2]{\vspace{0.4em}{\small\textsc{\color{#1}{#2}}}\par\vspace{0.1em}}
\usepackage{amssymb} 

\usepackage{colortbl}
\usepackage{tabularx}
\usepackage{xcolor}
\newcommand{\cellfavour}[1]{\textbf{#1}}

  \newcommand{\new}[1]{{#1}}
\usepackage{cmap}
\usepackage[T1]{fontenc}

\usepackage{enumitem}
\usepackage{amsmath}
\usepackage{microtype}
\usepackage{inconsolata}
\usepackage{xspace}

\newcommand{\ccfull}{\textsc{Conversation Coach}\xspace}

\title{Conversation Coach: A Voice-enabled AI System that Helps Practice Difficult Workplace Conversations}

\author{
  \textbf{Fanyou Wu$^{*}$ \quad Suraj Maharjan$^{*}$ \quad Ainur Yessenalina} \\
  \textbf{Dennis Xu Chen \quad Rahul Srivastava \quad Srinivasan H. Sengamedu} \\
  Amazon \\
  {\normalfont\small $^{*}$\,Equal contribution.}\\
  {\texttt{\{fanyouwu, mhjsuraj, yessenal, dennydc, rahsri, sengamed\}@amazon.com}}
}

\begin{document}
\maketitle
\begin{abstract}
Effective manager-employee communication is critical for retaining high performers and developing underperformers, yet training managers in these skills remains costly. Text-based chatbots offer a scalable approach but cannot provide realistic rehearsal: managers need to practice speaking aloud to build confidence before high-stakes conversations. In this paper, we propose \ccfull, a voice-first AI system that enables managers to rehearse difficult workplace conversations in a realistic spoken format. The system addresses three challenges: achieving low-latency interactions with strong language understanding, enabling adaptive conversations through configurable bot personalities that simulate different employee types, and generating personalized feedback on content and policy compliance. We compare an end-to-end speech-to-speech model with a cascaded approach combining automatic speech recognition, a large language model, and text-to-speech synthesis. The end-to-end approach achieves 3$\times$ lower median (P50) latency with native barge-in capability at an estimated 8$\times$ lower cost, while the cascaded approach offers superior reasoning essential for coaching quality. We deployed the cascaded architecture in production, where 40,000+ managers used it over six months, with adoption patterns indicating selective use for difficult conversations.
\end{abstract}

\section{Introduction}\label{sec:intro}

Effective workplace communication, such as delivering difficult feedback, navigating conflict, and conducting sensitive conversations, is a critical managerial skill with significant organizational impact~\cite{wilhelm2025commcoach}. Traditional training approaches such as workshops, role-play with human partners, and executive coaching are expensive, difficult to scale, and often unavailable when managers need to practice the most: immediately before a challenging conversation.

Voice-based AI coaching offers a promising alternative. Unlike text chat, voice interaction requires managers to practice speaking aloud in real time, creating a higher-fidelity rehearsal environment~\cite{macleod2010production}. However, building effective voice coaching systems requires navigating complex tradeoffs between response latency, language understanding quality, and conversational naturalness~\cite{ji2024wavchat}.

The architectural landscape for speech-to-speech AI presents a fundamental choice. End-to-end models such as Moshi~\cite{defossez2024moshi} process speech directly, offering lower latency and native handling of interruptions. Cascaded pipelines chain separate Automatic Speech Recognition (ASR), Large Language Model (LLM), and Text to Speech (TTS) components, providing modularity and access to state-of-the-art text reasoning but accumulating latency~\cite{paek2008automating}. Recent work suggests this tradeoff may be less stark than assumed~\cite{xtalk2025}, but production case studies comparing specific systems in coaching applications remain scarce. We make three contributions:

\begin{enumerate}[leftmargin=*,itemsep=2pt]
    \item \textbf{A domain-specific design framework}  organizing coaching-relevant tradeoffs across three dimensions to guide practitioners selecting between these approaches (\S\ref{sec:framework}).
    
    \item \textbf{A production system comparison} of Nova Sonic~2 (NS2; end-to-end) and a cascaded pipeline (Amazon Transcribe + Claude Sonnet 4.5 + Amazon Polly), characterizing their latency, cost, and coaching quality tradeoffs within this specific application (\S\ref{sec:experiments}).
    
    \item \textbf{An organization-wide deployment} of a voice-first coaching system to 40,000+ managers, with adoption patterns revealing selective, just-in-time use for difficult conversations and actionable engineering lessons (\S\ref{sec:deployment}, \S\ref{sec:discussion}).

\end{enumerate}

\begin{figure*}[!t]
\centering
  \includegraphics[width=\linewidth]{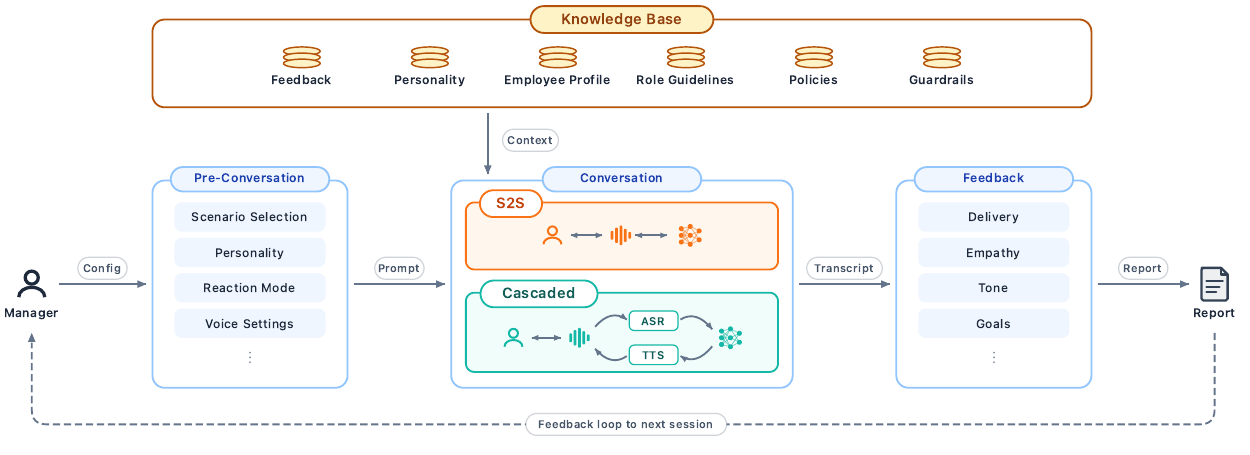}
  \caption{\ccfull System Design. The Pre-conversation module configures the simulation by allowing managers to select an employee and assign personality and reaction modes. The Conversation module offers two alternatives: Speech-to-Speech (end-to-end streaming with native barge-in) or Cascaded (ASR $\to$ LLM $\to$ TTS, each stage streaming to the next to minimize latency). The Feedback module provides actionable insights on content, framing, and policy compliance.}
  \label{fig:system}
\end{figure*}

\section{Related Work}\label{sec:related}

\vspace{\baselineskip}\noindent\textbf{Speech-to-Speech Dialogue Systems.} Spoken dialogue systems follow two paradigms: cascaded pipelines chaining ASR, LLM, and TTS components~\cite{paek2008automating}, and end-to-end models processing speech directly~\cite{borsos2023audiolm, zhang2023speechgpt, defossez2024moshi}. While end-to-end approaches promise lower latency and preservation of paralinguistic cues, recent analyses~\cite{xtalk2025, ji2024wavchat} suggest optimized cascaded systems can achieve competitive latency while retaining superior text-based reasoning. Our work contributes an empirical comparison of these paradigms in a deployed coaching application.

\vspace{\baselineskip}\noindent\textbf{AI-Based Coaching and Tutoring.}
Large language models have transformed conversational AI through instruction-following capabilities. Prior work includes spoken tutoring systems~\cite{litman2004itspoke}, large-scale deployed tutoring~\cite{afzal2019deployment}, AI-augmented executive coaching~\cite{arakawa2020inward}, deployed mental-health coaching~\cite{fitzpatrick2017woebot}, and recent studies on AI-assisted workplace communication training~\cite{wilhelm2025commcoach}. LLM-based persona simulation has been explored through role-playing frameworks~\cite{wang2024rolellm, zhou2024characterglm} that maintain character consistency across extended interactions, a challenge central to our system.

\vspace{\baselineskip}\noindent\textbf{Real-Time Conversational Dynamics.} Current spoken dialogue systems typically exhibit 700--1000ms latency~\cite{skantze2021turntaking, castillolopez2025turntaking}. Predictive turn-taking models~\cite{ekstedt2020turngpt, ekstedt2022vap} and emotional responsiveness~\cite{busso2008iemocap, poria2019meld} are active research areas relevant to coaching applications.

\section{Design Principles}
\label{sec:framework}

Building an effective voice-based coaching system requires addressing three interconnected challenges.

\vspace{\baselineskip}\noindent\textbf{Real-time spoken interaction.} Coaching demands responsiveness that feels natural~\cite{skantze2021turntaking}. \emph{Time to First Audio (TTFA)} measures the delay from when a user finishes speaking to when the system begins its response. End-to-end models avoid inter-component serialization; cascaded pipelines accumulate latency across stages, though streaming mitigates this. \emph{Barge-in support} allows users to interrupt mid-response, which is common in emotional conversations. End-to-end models handle this natively; cascaded approaches require explicit interruption detection.

\vspace{\baselineskip}\noindent\textbf{Adaptive persona simulation.} The bot must consistently portray a specific employee persona across an entire conversation. This demands strong instruction following (adhering to personality type, reaction mode, and guardrails) and contextual consistency (maintaining character even when the user's statements are surprising). Cascaded architectures benefit from state-of-the-art LLMs with strong instruction-following capabilities; end-to-end models must balance persona consistency against speech quality and latency.

\vspace{\baselineskip}\noindent\textbf{Feedback generation.} Post-conversation feedback analyzes \emph{what} the manager said (content: word choices, policy compliance, key talking points). Future extensions could integrate \emph{how} they said it (delivery: tone, pace, empathy, vocal confidence), where end-to-end models can potentially extract delivery features during conversation and cascaded pipelines would require a separate audio analysis module.
Table~\ref{tab:framework} summarizes how each approach addresses these challenges.

\begin{table}[!tbp]
\centering

\small
\setlength{\tabcolsep}{4pt}
\begin{tabular}{@{}lcc@{}}
\toprule
\textbf{Design Space} & \textbf{E2E} & \textbf{Cascaded} \\
\midrule
\multicolumn{3}{@{}l}{\textit{Real-time Interaction}} \\
\addlinespace[2pt]
\quad Time to first audio    & \cellfavour{Lower}   & Higher \\
\quad Barge-in support        & \cellfavour{Native}  & Requires logic \\
\addlinespace[4pt]
\multicolumn{3}{@{}l}{\textit{Conversational Quality}} \\
\addlinespace[2pt]
\quad Instruction following   & Moderate & \cellfavour{Strong} \\
\quad Persona consistency     & Moderate & \cellfavour{Strong} \\
\quad Content reasoning       & Moderate & \cellfavour{Strong} \\
\addlinespace[4pt]
\multicolumn{3}{@{}l}{\textit{System Engineering}} \\
\addlinespace[2pt]
\quad Delivery analysis       & Integrated & \cellfavour{Separate module} \\
\quad Component upgradability & Coupled    & \cellfavour{Independent} \\
\quad Debugging granularity   & Opaque     & \cellfavour{Per-component} \\
\bottomrule
\end{tabular}
\caption{\label{tab:framework} Qualitative architecture comparison based on production experience. E2E: end-to-end speech-to-speech (NS2); Cascaded: ASR + LLM + TTS pipeline (Transcribe + Claude + Polly).}
\end{table}

\section{System Architecture}
\label{sec:system}

We instantiate our tradeoff analysis in \ccfull, a production system for practicing difficult workplace conversations.
Figure~\ref{fig:system} shows the overall architecture, which comprises three modules aligned with the design spaces identified in \S\ref{sec:framework}.

\subsection{Scenario Configuration}
\label{sec:system-preconv}

A key design goal of the pre-conversation module is to generate system prompts that elicit realistic, varied employee behavior during role-play.
To this end, the module aggregates two types of context: \emph{scenario context} drawn from a knowledge base (company philosophy, policies, role-specific guidelines), and \emph{employee context} including performance history, contributions, peer and manager feedback, and self-reflection statements.
Managers configure each practice session by selecting a conversation scenario (e.g., annual review, promotion or compensation discussion), an employee profile, and behavioral parameters described below.

\vspace{\baselineskip}\noindent\textbf{Personality and Reaction Typology.}
Realistic coaching requires the simulated employee to exhibit a coherent behavioral profile rather than generic responses.
Drawing on Marston's DISC model~\cite{marston1928emotions} and Stone and Heen's taxonomy of feedback reception styles~\cite{stone2015thanks}, we define three personality types: Direct~/~Assertive, Passive~/~Conflict Avoidant, and Analytical~/~Detail Oriented. We consolidate the original four DISC quadrants into three types because the Influential and Steady styles both prioritize interpersonal harmony, yielding similar conversational behavior in our coaching context; separating them did not produce distinguishable role-play dynamics in pilot testing.

We cross these with three reaction modes that capture common emotional responses to negative feedback: Defensive~/~Deflective (challenging, justifying, or redirecting blame), Withdrawn (disengaging with minimal or flat responses), and Distressed (exhibiting emotional overwhelm such as raised voice or tearfulness).
The resulting $3 \times 3$ matrix yields nine personality-reaction combinations, providing systematic coverage of diverse conversational dynamics (see Appendix~\ref{appendix:reaction} for the full typology with illustrative utterances).

\subsection{Real-Time Role-Play}
\label{sec:system-conv}

The conversation module facilitates real-time spoken role-play between the manager and the simulated employee. We implement both architectural approaches to enable direct comparison under identical coaching scenarios.

\vspace{\baselineskip}\noindent\textbf{End-to-End Approach (Nova Sonic~2; NS2).}
Nova Sonic~2~\cite{aws2025novasonic2} processes audio input and generates audio output within a single model, handling speech recognition, language understanding, response generation, and speech synthesis without inter-component serialization.
This achieves faster TTFA with native barge-in support and preservation of paralinguistic cues from input audio.

\vspace{\baselineskip}\noindent\textbf{Cascaded Approach.}
Amazon Transcribe~\cite{aws2025transcribe} (ASR), Claude Sonnet 4.5~\cite{anthropic2025claude} (response generation), and Amazon Polly~\cite{aws2025polly} (TTS) prioritize reasoning quality over latency.
Streaming is applied at each stage to mitigate accumulated latency; without it, end-to-end delay makes the system unusable (see \S\ref{sec:discussion} for measurements).
For barge-in handling, the client monitors the ASR stream for new speech onset during TTS playback; upon detection, TTS audio is immediately halted and the new user utterance is processed as a fresh turn.
The assistant turn in conversation history is truncated to the portion delivered before the interruption.
This approach requires explicit client-side detection and playback control, unlike NS2's native interruption handling.

\subsection{Post-Conversation Feedback}
\label{sec:system-feedback}

The feedback module focuses on \emph{content analysis}: word choices, message framing, policy compliance, and key talking points.
A critical function is detecting privacy violations such as revealing confidential calibration ratings.
Claude synthesizes the analysis into structured feedback: a conversation summary, specific strengths, and areas for improvement with actionable rephrasing suggestions (see Appendix~\ref{appendix:feedback} for the prompt).
The system stores all conversation history and feedback, enabling managers to revisit sessions and practice the same scenario across different personality--reaction combinations.

\section{Evaluation}
\label{sec:experiments}

We evaluate whether the architectural tradeoffs identified in \S\ref{sec:framework} manifest in practice, focusing on latency, persona consistency, speech quality, and coaching feedback quality.

\subsection{Dataset}
\label{sec:dataset}

We constructed an evaluation dataset in two stages.
First, 10 human-conducted conversations established a ground-truth rubric for validating LLM judge reliability (\S\ref{sec:results}).
We then created 148 synthetic scenarios spanning all 27 configurations (3 employee profiles $\times$ 9 personality-reaction types, ${\sim}$6 per configuration). Each profile contains role description, performance history, self-reflection statements, peer feedback, and manager feedback.
Manager turns were generated by an LLM and synthesized to speech via TTS, so both architectures process realistic audio input through their full pipelines.
Each scenario was replayed identically to both systems, yielding 296 transcripts (148 paired comparisons). A human reviewer excluded 14 of the initial 162 sessions for off-topic drift or premature termination.

\subsection{Experimental Setup}
\label{sec:setup}

We use Nova Sonic~2 (\texttt{amazon.nova-2-sonic-v1:0}, voice \texttt{tiffany}) for the end-to-end model and, for the cascaded approach, Amazon Transcribe in streaming mode for ASR, Claude Sonnet~4.5 (\texttt{anthropic.claude-sonnet-4-5-20250929-v1:0}) via Amazon Bedrock for response generation, and Amazon Polly (generative engine, voice \texttt{Ruth}) for TTS.
All LLM operations use temperature~$= 0.7$.
Speech quality is assessed using UTMOS~\cite{saeki2022utmos}.

\subsection{Results}
\label{sec:results}

We compare NS2 and the cascaded approach across latency, persona consistency, speech quality, and coaching feedback (Table~\ref{tab:results}).

\begin{table}[!ht]
\centering
\small
\setlength{\tabcolsep}{4pt}
\renewcommand{\arraystretch}{1.15}
\begin{tabular}{@{}llc@{\hspace{4pt}}}
\toprule
\textbf{Metric} & \textbf{Gap} & \textbf{Favors} \\
\midrule
\multicolumn{3}{@{}l}{\textit{LLM Judge Evaluation}} \\
\addlinespace[2pt]
\quad Initial Reaction & marginal & -- \\
\quad Stakeholder Response & marginal & -- \\
\quad Authority Response & marginal & -- \\
\quad Feedback Reception & mixed & Cascaded \\
\quad Timeline Pushback & marginal & -- \\
\quad Directive Compliance & substantial & Cascaded \\
\quad Language Pattern & substantial & Cascaded \\
\quad Role Adherence & substantial & Cascaded \\
\quad Authenticity & substantial & Cascaded \\
\addlinespace[4pt]
\multicolumn{3}{@{}l}{\textit{Other Metrics}} \\
\addlinespace[2pt]
\quad UTMOS~\cite{saeki2022utmos} & 4.43 vs 4.41 & -- \\
\quad P50 TTFA (s) & 1.4 vs 4.2 & NS2 \\
\quad P99 TTFA (s) & 2.8 vs 7.6 & NS2 \\
\quad TM99 TTFA (s) & 1.5 vs 7.0 & NS2 \\
\quad Est.\ cost (\$/session) & \$0.05 vs \$0.39 & NS2 \\
\bottomrule
\end{tabular}
\caption{\label{tab:results} Comparison of Nova Sonic~2 (NS2) and the cascaded pipeline across behavioral and speech quality metrics. LLM judge results are majority votes across four models (Table~\ref{tab:multi-judge}). Gap = mean score difference on a 5-point scale: \textit{marginal}\,($<$0.5), \textit{moderate}\,(0.5--1.5), \textit{substantial}\,($>$1.5). \textit{Mixed} = 2 judges marginal, 2 moderate. TM99 = trimmed mean excluding values above the 99th percentile. Per-judge breakdowns in Table~\ref{tab:multi-judge}.}
\end{table}

\vspace{\baselineskip}\noindent\textbf{Latency.}
Measured server-side across all conversation turns, the median TTFA (P50) for NS2 is 1.4s compared to 4.2s for the cascaded approach (3$\times$ improvement). At the 99th percentile (P99), NS2 remains responsive at 2.8s while the cascaded pipeline reaches 7.6s; the trimmed mean excluding values above P99 (TM99) confirms this gap is consistent (1.5s vs 7.0s), not driven by outliers.
Both approaches use streaming throughout; despite this optimization, the cascaded pipeline's sequential processing results in noticeably higher first-audio latency.

\vspace{\baselineskip}\noindent\textbf{ASR Accuracy.}
On the manager turns used in our evaluation (Polly-synthesized speech with known ground-truth text), Amazon Transcribe achieves 5.1\% word error rate (WER). A matched per-turn measurement for NS2 was not possible, as its streaming transcript is not segmented by turn. For external reference, Transcribe reports ${\sim}5.6\%$ WER in streaming mode~\cite{picovoice2024benchmark} and NS2 ${\sim}6.5\%$ on CommonVoice~\cite{aws2025novasonic2}; although these benchmarks use differing test conditions, the similar ranges suggest ASR quality does not meaningfully differentiate the two approaches in our setting.

\vspace{\baselineskip}\noindent\textbf{Persona Consistency and Reasoning.}
LLM judges evaluated all conversations spanning all nine personality-reaction combinations, assessing whether the bot (1)~maintained its employee role throughout, (2)~followed the assigned personality and reaction mode, (3)~pushed back appropriately, and (4)~challenged feedback rather than passively accepting it.

Both approaches show marginal differences on conversational dimensions (initial reaction, stakeholder response, authority response), indicating comparable performance on simpler interaction patterns.
The cascaded approach shows a mixed result on feedback reception (split across judges) and substantial advantages on persona consistency dimensions, particularly in maintaining complex behavioral instructions across extended conversations. The cascaded advantage on role adherence and authenticity is unanimous across all four judge models; directive compliance and language pattern show 3-of-4 agreement.
NS2 exhibits occasional character breaks under complex persona requirements but remains effective for straightforward scenarios.

\new{\vspace{\baselineskip}\noindent\textbf{Isolating Architecture from Model.}
Because the cascaded pipeline pairs a specific LLM (Claude Sonnet~4.5) with the cascaded architecture, its persona advantage could in principle stem from the model rather than the architecture. To disentangle the two, we re-ran the cascaded pipeline with a substantially weaker and cheaper LLM, Claude Haiku~4.5, on the same scenarios and four-judge panel. A two one-sided tests (TOST) equivalence analysis (equivalence margin $\pm 1$ Likert step) finds Sonnet and Haiku statistically equivalent on 8 of 9 persona dimensions, all at $p < 0.01$ across all four judges; only Initial Reaction falls outside the equivalence margin. Since a much weaker LLM preserves the persona advantage over NS2, the gap is attributable to the cascaded architecture's ability to leverage any adequate text-based LLM through explicit system prompting, rather than to Sonnet's specific instruction-following strength.

\vspace{\baselineskip}\noindent\textbf{Human--LLM Agreement.}
To validate judge reliability, three annotators independently scored 54 conversations (27 per system, spanning all configuration cells) on six persona-relevant dimensions (Initial Reaction, Feedback Reception, Stakeholder Response, Language Pattern, Role Adherence, and Authenticity) applicable across all personality types. Annotators were blinded to system identity: transcripts were presented in randomized order with system names removed. Human--LLM agreement reaches Cohen's $\kappa = 0.81$, exceeding the human--human agreement of $\kappa = 0.67$; that is, the LLM judge agrees with each annotator more closely than the annotators agree with one another, confirming that the judge produces reliable scores across all configurations.

\vspace{\baselineskip}\noindent\textbf{Inter-Model Agreement.}
We repeated the evaluation with four diverse judge models (Claude, Nova, MiniMax, and GLM): Fleiss' $\kappa = 0.41$ (moderate) with 83.3\% majority consensus. The cascaded advantage on Role Adherence and Authenticity is unanimous across all four judges, while Directive Compliance and Language Pattern show 3-of-4 agreement (the Nova judge rates these as marginal; Table~\ref{tab:multi-judge}). Potential same-family bias (Claude judge favoring the Claude-powered cascaded system; Nova judge favoring NS2) is mitigated by the inclusion of two independent judges (MiniMax, GLM) with no family overlap, both of which align with the majority on all unanimous dimensions.}

\vspace{\baselineskip}\noindent\textbf{Speech Quality.}
After filtering responses shorter than 3 seconds (which are primarily barge-in fragments), UTMOS~\cite{saeki2022utmos} evaluation (scale 1--5) shows both architectures produce high-quality speech.

\vspace{\baselineskip}\noindent\textbf{Cost.} For a typical 10-turn session, we estimate the cascaded approach costs approximately \$0.39 (Transcribe \$0.02, Claude \$0.19, Polly \$0.18) versus \$0.05 for NS2, an 8$\times$ reduction. The primary cost drivers are generative TTS and the LLM's growing context window across turns.

\vspace{\baselineskip}\noindent\textbf{Coaching Feedback.}
Of the same 148 synthetic conversations, 28 contain intentional policy violations (managers disclosing confidential calibration ratings) and the rest do not, providing known ground-truth labels for evaluating violation detection. To mitigate model-family circularity, we evaluate detection with four diverse classifiers (Claude Opus~4.5, Nova Pro, MiniMax M2.5, GLM-5; identifiers in Table~\ref{tab:judge-models}). All four achieve perfect recall and high precision, yielding F1 scores of 0.92--0.97 (Table~\ref{tab:violation-detection}). False positives across all models were over-caution: flagging indirect references (e.g., ``you are in the solid performer category''), which is preferable in a compliance-sensitive context.

\begin{table}[tb]
\centering
\small
\begin{tabular}{@{}lccc@{}}
\toprule
\textbf{Model} & \textbf{Precision} & \textbf{Recall} & \textbf{F1} \\
\midrule
Claude Opus~4.5 & 0.93 & 1.00 & 0.97 \\
Nova Pro & 0.85 & 1.00 & 0.92 \\
MiniMax M2.5 & 0.90 & 1.00 & 0.95 \\
GLM-5 & 0.90 & 1.00 & 0.95 \\
\bottomrule
\end{tabular}
\small
\caption{\label{tab:violation-detection} Confidentiality-violation detection across four diverse classifier models, evaluated against 148 ground-truth labels (28 violations, 120 non-violations). All models achieve perfect recall; false positives are over-cautious flags of indirect references.}
\end{table}

\new{\section{Deployment and Usage Patterns}
\label{sec:deployment}

To assess real-world viability beyond offline evaluation, we report deployment metrics from an organization-wide launch available to all managers during the annual review cycle covering a six-month period.

Based on the evaluation results in \S\ref{sec:experiments}, we deployed the \textbf{cascaded architecture} in production, prioritizing persona consistency and reasoning quality over latency. The 22\% return rate (\S\ref{sec:deploy-adoption}) and stable session durations suggest that the cascaded latency was acceptable for this coaching context, though we cannot rule out that lower latency would have driven higher engagement without an A/B comparison.
Table~\ref{tab:deployment-summary} summarizes key adoption figures.

\begin{table}[!h]
\centering
\small
\begin{tabular}{lr}
\toprule
\textbf{Metric} & \textbf{Value} \\
\midrule
Total sessions & 54,000+ \\
Unique managers & 40,000+ \\
Returning users & 22\% \\
Median session duration & 135\,s \\
\bottomrule
\end{tabular}
\small
\caption{\label{tab:deployment-summary} Deployment summary over a six-month period.}
\end{table}

\subsection{Adoption Patterns}
\label{sec:deploy-adoption}

Adoption is notably higher among newer front-line managers (over 3$\times$ the overall adoption rate across all eligible managers), and the 22\% return rate and median 135\,s session duration confirm substantive engagement rather than superficial exploration.
Technical managers engage at higher practice rates than non-technical managers, while newer managers spend 19\% more time per session than tenured peers.
Usage peaks 4$\times$ above the period-average session rate in the two weeks preceding review deadlines, confirming just-in-time preparation rather than general skill development.
Among sessions that begin practice, 36\% complete the full voice conversation and 23\% of completers request feedback, indicating progressive commitment once the initial activation barrier is overcome.
Qualitative feedback from active users corroborates the engagement metrics. One manager noted: \emph{``The roleplay functionality works well, picking up my varied tone and inflection. It also sounded like a true conversation.''} Another shared: \emph{``The role play functionality successfully demonstrated a challenging performance review scenario.''}

\subsection{Selective Use for Difficult Conversations}
\label{sec:deploy-targeting}

Managers disproportionately use the system for their most challenging conversations: employees in the lowest performance tiers appear in 43.3\% of unique coached manager-employee pairs (nearly 2$\times$ their baseline rate), and managers prepare for only 17.6\% of conversations on average, with 84\% preparing for fewer than 25\% of their direct reports, indicating deliberate, targeted use rather than routine application.
Within-team comparisons are consistent with this selection effect: coached employees exhibit 10.9 percentage points (pp) lower career growth sentiment (a survey-based measure of perceived development opportunities) than non-coached teammates prior to the conversation, reflecting pre-existing challenges that prompted the manager to seek practice.

\subsection{Preliminary Outcome Analysis}
\label{sec:deploy-outcomes}
We compared coached employees who experienced a rating decline to matched controls whose managers did not use the system, using a doubly robust estimator~\cite{bang2005doubly} that combines propensity weighting with outcome regression and adjusts for tenure, level, job family, location, and prior rating. The current sample is underpowered (minimum detectable effect, MDE $\approx 13$ percentage points, pp); the observed 4.3\,pp difference in job satisfaction between coached and control groups is not statistically significant and should be read only as motivation for a future adequately powered study.}

\section{Discussion}
\label{sec:discussion}

The cascaded pipeline's persona advantage is architectural rather than a byproduct of the specific LLM: substituting a substantially weaker model (Claude Haiku~4.5) preserves the advantage over NS2 (\S\ref{sec:results}). These findings are nonetheless specific to the systems evaluated; other instantiations may shift the tradeoff boundary. Beyond this comparison, our deployment surfaced several engineering lessons.

\vspace{\baselineskip}\noindent\textbf{Streaming is non-negotiable.} Without end-to-end streaming across ASR, LLM, and TTS, accumulated latency exceeds 8 seconds. Only full-pipeline streaming, where TTS begins synthesizing as soon as the first LLM tokens arrive, achieves acceptable conversational flow.

\vspace{\baselineskip}\noindent\textbf{Prompt sensitivity.} NS2 occasionally simplifies complex persona instructions, particularly when multiple behavioral constraints interact. Claude Sonnet 4.5 follows detailed prompts reliably but requires engineering to avoid verbose, text-style outputs; we constrain response length and use Speech Synthesis Markup Language (SSML) directives to produce natural speech patterns.

\vspace{\baselineskip}\noindent\textbf{Persona drift.} In longer conversations (15+ turns), both architectures gradually weaken adherence to the assigned personality; the cascaded approach is more robust, likely due to stronger context-window utilization.

\vspace{\baselineskip}\noindent\textbf{Choosing an architecture.}
Drawing on our deployment experience, we offer practitioner guidance. \emph{Choose end-to-end} when sub-2-second latency is essential, the application requires frequent natural interruptions, persona complexity is moderate, or paralinguistic preservation is prioritized over post-hoc analysis. \emph{Choose cascaded} when complex instruction following and persona consistency are critical, reasoning about conversation context is needed, or component-level debugging and independent upgradability matter and the added (streaming-mitigated) latency is acceptable. \emph{Consider hybrid} designs that combine both, e.g., an end-to-end model for real-time conversation while routing transcripts to a cascaded pipeline for reasoning.

\section{Conclusion}\label{sec:conclusion}

We presented \ccfull, a voice-first AI coaching system that enables managers to rehearse difficult conversations in a realistic spoken format.
Our comparison of NS2 and a cascaded pipeline (Transcribe + Claude Sonnet~4.5 + Polly) reveals a tradeoff: NS2 achieves 3$\times$ lower latency and 8$\times$ lower cost with native barge-in, while the cascaded pipeline offers superior persona consistency for complex coaching scenarios. These findings are specific to the evaluated systems; the tradeoff frontier may shift as both paradigms evolve.
Deployment across tens of thousands of managers shows adoption patterns consistent with selective use for the most challenging conversations, though adoption alone does not establish coaching effectiveness. Future work includes audio-based delivery feedback, hybrid architectures, and extending beyond manager-focused scenarios.

\section*{Limitations}
\label{sec:limitations}

\new{Our work has several limitations.
(1)~\textbf{No causal proof of downstream effectiveness:} we have not established that simulated practice improves real-world conversations; the current treated sample is underpowered and a longitudinal controlled study remains essential future work.
(2)~\textbf{Synthetic data and LLM judges:} the architecture comparison relies on 148 LLM-generated conversations scored by LLM judges, validated against human annotators only on a subset (\S\ref{sec:results}). Data-sharing constraints further limit us to relative gap levels rather than absolute per-system scores.
(3)~\textbf{No delivery feedback:} the current system analyzes only content (word choices, policy compliance); delivery analysis (tone, pace, empathy) from audio features remains future work.
(4)~\textbf{Proprietary models:} we evaluated only Claude Sonnet~4.5 and NS2; generalization to open-source alternatives~\cite{meta2025llama4, qwen3} is unexplored, though the modular architecture allows substitution.

}

\section*{Ethical Considerations}
\label{sec:ethics}

The system could reinforce particular management styles; we mitigate this through configurable persona parameters.
Practice sessions are visible only to the individual manager to protect against exposing communication weaknesses.
To comply with privacy regulations, no audio recordings are stored; conversation transcripts are retained for a maximum of 30 days and then permanently deleted.
We position the system as a complement to, not replacement for, human coaching.
The simulated personas do not capture the full range of human responses, particularly those informed by cultural or identity-related communication norms.

\bibliography{reference}

\appendix
\onecolumn

\section{Multi-Model LLM-as-a-Judge}\label{appendix:multi-judge}

\begin{table*}[h]
\centering

\begin{subtable}[t]{\textwidth}
\centering
\small

\begin{tabular}{@{}ll@{}}
\toprule
\textbf{Model} & \textbf{Model ID} \\
\midrule
Claude Opus~4.5~\cite{anthropic2025claude} & \texttt{claude-opus-4-5-20251101-v1:0} \\
Nova Pro~\cite{aws2025novapro} & \texttt{us.amazon.nova-pro-v1:0} \\
MiniMax M2.5~\cite{minimax2025m25} & \texttt{minimax.minimax-m2.5} \\
GLM-5~\cite{zhipu2025glm} & \texttt{zai.glm-5} \\
\bottomrule
\end{tabular}
\caption{Judge models.}\label{tab:judge-models}
\end{subtable}

\vspace{8pt}

\begin{subtable}[t]{\textwidth}
\centering
\small
\setlength{\tabcolsep}{4pt}

\begin{tabular}{@{}lcccc@{}}
\toprule
\textbf{Dimension} & \textbf{Claude Opus 4.5} & \textbf{Nova Pro} & \textbf{MiniMax M2.5} & \textbf{GLM-5} \\
\midrule
\multicolumn{5}{@{}l}{\textit{Marginal Gap (Comparable)}} \\
\addlinespace[2pt]
\quad Initial Reaction      & marginal & marginal & marginal & marginal \\
\quad Stakeholder Response  & marginal & marginal & marginal & marginal \\
\addlinespace[4pt]
\multicolumn{5}{@{}l}{\textit{Mixed Results}} \\
\addlinespace[2pt]
\quad Feedback Reception    & moderate & moderate & marginal & marginal \\
\quad Timeline Pushback     & marginal & moderate & marginal & marginal \\
\quad Authority Response    & marginal & marginal & marginal & moderate \\
\addlinespace[4pt]
\multicolumn{5}{@{}l}{\textit{Substantial Gap (Cascaded Advantage)}} \\
\addlinespace[2pt]
\quad Directive Compliance  & substantial & marginal & substantial & substantial \\
\quad Language Pattern      & substantial & marginal & substantial & substantial \\
\quad Role Adherence        & substantial & substantial & substantial & substantial \\
\quad Authenticity          & substantial & substantial & substantial & substantial \\
\bottomrule
\end{tabular}
\caption{Per-dimension gap levels by judge model.}\label{tab:multi-judge}
\end{subtable}
\caption{\label{tab:multi-judge-eval} Multi-model LLM-as-a-Judge evaluation. \textbf{(a)} The four judge models. \textbf{(b)} Per-dimension gap levels for each judge, on the same scale as Table~\ref{tab:results} (mean score difference on a 5-point scale: marginal $<$0.5, moderate 0.5--1.5, substantial $>$1.5). Inter-model agreement: Fleiss' $\kappa = 0.41$, majority consensus 83.3\%.}
\end{table*}

\section{Employee Reaction Typology}\label{appendix:reaction}
\begin{table*}[!h]
\centering
\vspace{1ex}
\small
\resizebox{0.95\textwidth}{!}{%
\begin{tabular}{p{2.2cm} p{4.5cm} p{4.5cm} p{4.5cm}}
\toprule
\textbf{Reaction Mode} & \textbf{Direct / Assertive} & \textbf{Passive / Conflict Avoidant} & \textbf{Analytical / Detail Oriented} \\
\midrule
\textbf{Defensive /\newline Deflective}
& Challenges manager's credibility or motives (e.g., ``You're only bringing this up now because you don't like me speaking up in meetings'').
Interrupts with justifications (e.g., ``But what about the Johnson project? I delivered that ahead of schedule'').
Directly confronts manager's assessment. Voice may be firm and confrontational.
& Makes indirect excuses (e.g., ``I think there might have been some confusion about the deadlines\ldots'').
Deflects by bringing up others' issues subtly (e.g., ``It's just that others on the team also miss deadlines sometimes\ldots'').
Agrees superficially while making excuses. Avoids direct accountability, gently shifts blame. Voice is uncertain and soft.
& Defends self with data, points out inconsistencies (e.g., ``According to the guidelines, I'm actually exceeding the role level expectations'').
Voice may sound cold or formal. \\
\midrule
\textbf{Withdrawn}
& Gives clipped, non-committal responses (e.g., ``Sure. Whatever you say'').
Tone is flat or dismissive.
& Minimal speech with delayed or one-word responses (e.g., ``okay'').
& Withdrawn into logic, gives only data-based answers (e.g., ``As per last quarter's report, the error rate was 1.2\%''). Emotionally flat. \\
\midrule
\textbf{Distressed}
& Appears overwhelmed, raises voice, may interrupt (e.g., ``I'm doing everything I can and you still say I'm not doing enough?'').
& Tearful or trembling voice, sounds defeated (e.g., ``I don't think I can handle this anymore'').
& Voice trembling due to ambiguity of the feedback (e.g., ``This feels subjective; I don't know how to fix something I can't measure''). \\
\bottomrule
\end{tabular}
}
\caption{\label{tab:emotional-reactions} Typology of employee emotional reactions to negative performance feedback, organized by reaction mode and personality type. Example utterances are illustrative.}
\end{table*}

\section{Prompt}\label{appendix:prompt}
\subsection{Conversation Prompt}\label{appendix:roleplay}
\begin{promptbox}[title=Role-Playing System Prompt: Condensed prompt for employee simulation]{promptblue}
\scriptsize
 You are an XXX company employee named \texttt{\{role\_player\}} participating in an Annual Review conversation with your manager, \texttt{\{user\_name\}}.

\promptlabel{promptblue}{Character Parameters}
\begin{itemize}[noitemsep,topsep=0pt]
    \item Personality Type: \texttt{\{personality\}} (Direct/Assertive, Passive/Conflict Avoidant, or Analytical/Detail Oriented)
    \item Reaction Mode: \texttt{\{reaction\}} (Defensive/Deflective, Withdrawn, or Distressed)
\end{itemize}
Maintain these traits consistently. Do not break character under any circumstances.

\promptlabel{promptblue}{Conversation Flow}
Meeting acknowledgment $\rightarrow$ Rating inquiry $\rightarrow$ Feedback discussion $\rightarrow$ Self-reflection review $\rightarrow$ Peer feedback integration $\rightarrow$ Growth areas identification $\rightarrow$ Development planning $\rightarrow$ Career pathway discussion $\rightarrow$ Action item alignment

\promptlabel{promptblue}{Rating Systems Context}
\ldots

\promptlabel{promptblue}{Rating Reception Guidelines}
\begin{itemize}[noitemsep,topsep=0pt]
    \item Display realistic emotions (curiosity, apprehension, nervousness) when receiving ratings
    \item Compare received rating with self-assessment; express surprise if discrepancy exists
    \item Ask clarifying questions before discussing future growth
    \item Push back respectfully if ratings do not align with self-perception
     \item \ldots
\end{itemize}

\promptlabel{promptblue}{Response Guidelines}
\begin{itemize}[noitemsep,topsep=0pt]
    \item Keep responses concise (1-2 sentences) while maintaining character authenticity
    \item Express reactions consistent with assigned personality and reaction mode
    \item Ask specific questions about evaluation criteria and development metrics
    \item \ldots
\end{itemize}

\promptlabel{promptblue}{Guardrails}
\ldots

\promptlabel{promptblue}{Output Structure}
\begin{codeblock}
<speak>Spoken dialogue with basic punctuation only</speak>
<other>
[THOUGHT]: Internal analysis based on personality
[FEELING]: Emotional state aligned with reaction mode
</other>
\end{codeblock}

\promptlabel{promptblue}{Inputs}
\texttt{\{employee\}}, \texttt{\{employee\_self\_reflection\}}, \texttt{\{manager\_feedback\}}, \texttt{\{peer\_feedback\}}, \texttt{\{rating\_provided\_by\_manager\}}, \texttt{\{role\_guideline\}}, \texttt{\{personality\_reaction\_characteristics\}}
\end{promptbox}

\vspace{1em}

\clearpage
\subsection{Feedback Generation Prompt}\label{appendix:feedback}
\begin{promptbox}[title=Feedback Generation Prompt: Post-conversation coaching feedback]{promptgreen}
\scriptsize
 You are an AI voice coaching assistant providing real-time feedback to a manager after their annual review conversation practice. Speak directly to the manager in a conversational, supportive tone. Keep feedback concise and actionable.

\promptlabel{promptgreen}{Critical Restrictions}
\begin{itemize}[noitemsep,topsep=0pt]
    \item Do NOT discuss \ldots
    \item Focus only on coaching conversation skills and techniques for annual review discussions
    \item Emphasize developmental feedback delivery, goal alignment, and career development planning
\end{itemize}

\promptlabel{promptgreen}{Rating Systems Context}
 \ldots

\promptlabel{promptgreen}{Response Structure}
\begin{enumerate}[noitemsep,topsep=0pt]
    \item \texttt{\#\# Summary} - Overall effectiveness, key strength, main improvement area, one miss
    \item \texttt{\#\# What You Did Well} - Effective feedback delivery, emotional intelligence, communication clarity
    \item \texttt{\#\# What You Could Have Done Better} - Missed opportunities, unclear framing, unclear goals
\end{enumerate}

\promptlabel{promptgreen}{Context Assessment}
If fewer than 3 meaningful exchanges occurred, return insufficient context response. Look for: performance acknowledgment, development discussion, employee reactions, feedback delivery approach, concrete next steps.

\promptlabel{promptgreen}{Voice Tone Guidelines}
\begin{itemize}[noitemsep,topsep=0pt]
    \item Sound like a supportive coach, not a critic
    \item Use direct "you" language with short, conversational sentences
    \item Comment on tone matching importance of performance discussions
    \item Note sentiment alignment with annual review content
    \item End on an encouraging note
\end{itemize}

\promptlabel{promptgreen}{Inputs}
\texttt{\{talking\_points\}}, \texttt{\{chat\_history\}}
\end{promptbox}

\vspace{1em}

\subsection{Policy Violation Detection Prompt}\label{appendix:prompt-violation}
\begin{promptbox}[title=Policy Violation Detection Prompt: Assessing confidentiality compliance]{promptorange}
\scriptsize
 You are an AI system designed to evaluate whether confidentiality violations occurred in manager-employee conversations and whether the feedback system properly detected them.

\promptlabel{promptorange}{Task}
Analyze a conversation and its generated feedback to determine:
\begin{enumerate}[noitemsep,topsep=0pt]
\item whether the manager violated confidentiality rules, and
\item whether the feedback system properly flagged any violations.
\end{enumerate}

\promptlabel{promptorange}{Violation Rules}
\begin{itemize}[noitemsep,topsep=0pt]
    \item Managers must NOT disclose internal calibration ratings. Discussing only public rating is permitted. \ldots
\end{itemize}

\promptlabel{promptorange}{Verdict Logic}
\begin{itemize}[noitemsep,topsep=0pt]
    \item \textit{True}: Violation detected and flagged, or no violation and no false flag
    \item \textit{False}: Violation missed, or false positive flagging
\end{itemize}

\promptlabel{promptorange}{Output}
\begin{codeblock}
\{"violation\_in\_chat": bool,
 "flagged\_in\_feedback": bool,
 "reasoning": str,
 "final\_verdict": bool\}
\end{codeblock}

\promptlabel{promptorange}{Inputs}
\texttt{\{conversation\}}, \texttt{\{feedback\}}
\end{promptbox}

\vspace{1em}

\clearpage
\subsection{Evaluation Prompt}\label{appendix:evaluation}
\begin{promptbox}[title=LLM-as-a-Judge Prompt: Evaluating role-play quality]{promptpurple}
\scriptsize
You are an expert evaluator assessing the quality of an AI assistant's role-playing performance in a simulated annual review conversation. The assistant is playing the role of an employee receiving feedback from their manager.

\promptlabel{promptpurple}{Context}
\begin{itemize}[noitemsep,topsep=0pt]
    \item \textbf{Persona}: \{persona\}
    \item \textbf{Communication Style}: \{style\}
    \item \textbf{Emotional State}: \{emotion\}
\end{itemize}

The assistant should embody these characteristics throughout the conversation while responding to the manager's feedback delivery.

\promptlabel{promptpurple}{Conversation to Evaluate}
\texttt{\{conversation\}}

\promptlabel{promptpurple}{Evaluation Categories}
For each category, provide: (1) a score from 1--5 and (2) a brief justification (1--2 sentences).

\begin{enumerate}[noitemsep,topsep=0pt]
\item \textbf{Initial Reaction Assessment}: How well does the assistant's initial response reflect the assigned persona and emotional state? Does the reaction feel authentic to someone receiving performance feedback?
\item \textbf{Feedback Reception Analysis}: How appropriately does the assistant receive and process feedback? Does the response pattern match the assigned communication style (analytical/direct/passive)?
\item \textbf{Stakeholder Feedback Response}: How well does the assistant handle mentions of stakeholder/peer feedback? Does the response show appropriate engagement with external perspectives?
\item \textbf{Timeline and Expectations Pushback}: When timeline or expectations are mentioned, does the assistant appropriately push back or accept based on their persona? Is the level of challenge appropriate?
\item \textbf{Authority and Restrictions Response}: How does the assistant respond to authority figures and restrictions? Does the response reflect the assigned emotional state (defensive/distressed/withdrawn)?
\item \textbf{Directive Compliance Assessment}: Does the assistant follow conversational directives while maintaining character? Balance between compliance and persona authenticity?
\item \textbf{Language Pattern Analysis}: Does the assistant's language (word choice, sentence structure, tone) consistently reflect the assigned persona and style throughout?
\item \textbf{Role Adherence Assessment}: Does the assistant stay in character as an employee throughout? No breaks in role-play or inappropriate assistant-like responses?
\item \textbf{Authenticity Consistency}: Is the persona portrayal consistent from start to finish? No sudden personality shifts or contradictory behaviors?
\end{enumerate}

\promptlabel{promptpurple}{Output Format}
\begin{codeblock}
\{"evaluations": \{
    "Initial Reaction Assessment": \{"score": <1-5>, "justification": "<text>"\},
    "Feedback Reception Analysis": \{"score": <1-5>, "justification": "<text>"\},
    "Stakeholder Feedback Response": \{"score": <1-5>, "justification": "<text>"\},
    "Timeline and Expectations Pushback": \{"score": <1-5>, "justification": "<text>"\},
    "Authority and Restrictions Response": \{"score": <1-5>, "justification": "<text>"\},
    "Directive Compliance Assessment": \{"score": <1-5>, "justification": "<text>"\},
    "Language Pattern Analysis": \{"score": <1-5>, "justification": "<text>"\},
    "Role Adherence Assessment": \{"score": <1-5>, "justification": "<text>"\},
    "Authenticity Consistency": \{"score": <1-5>, "justification": "<text>"\}
  \},
  "overall\_score": <average of all scores>,
  "summary": "<2-3 sentence overall assessment>"\}
\end{codeblock}

\end{promptbox}

\end{document}